\documentclass[10pt,twocolumn,twoside]{IEEEtran}

\ifCLASSINFOpdf
   \usepackage[pdftex]{graphicx}
   \graphicspath{{Figures/}}
   \DeclareGraphicsExtensions{.eps,.pdf,.png}
   \usepackage{epstopdf}
	\usepackage{pdfpages}
\else
   \usepackage[dvips]{graphicx}
   \graphicspath{{Figures/}}
   \DeclareGraphicsExtensions{.eps}
\fi

\usepackage[cmex10]{amsmath}
\usepackage{algorithm}
\usepackage{algpseudocode}

\usepackage{array}
\usepackage{mdwmath}
\usepackage{mdwtab}

\usepackage{url}

\usepackage{amssymb,dsfont}
\usepackage{color}
\usepackage{ulem}

\def \bY{\mathbf{Y}}
\def \bS{\mathbf{S}}
\def \bA{\mathbf{A}}
\def \b0{\mathbf{0}}

\newcommand{\bs}[1]{\mathbf{#1}}

\begin{document}

\title{Sparse and Non-negative BSS for Noisy Data}

\author{J{\'e}r{\'e}my~Rapin,
        J{\'e}r{\^o}me~Bobin,
        Anthony~Larue,
        and~Jean-Luc~Starck\\firstname.lastname@cea.fr
\thanks{Copyright (c) 2013 IEEE. Personal use of this material is permitted. However, permission to use this material for any other purposes must be obtained from the IEEE by sending a request to pubs-permissions@ieee.org.}%
\thanks{Manuscript accepted on August $11^\text{th}$, 2013.}%
\thanks{A. Larue and J. Rapin are with CEA, LIST, 91191 Gif-sur-Yvette Cedex, France.}%
\thanks{J. Bobin, J.Rapin and J.-L. Starck are with CEA, IRFU, Service d'Astrophysique, 91191 Gif-sur-Yvette Cedex, France.}}


\markboth{IEEE Signal Processing,~2013}{Rapin \MakeLowercase{\textit{et al.\@}}: Sparse and Non-negative BSS for Noisy Data}


\maketitle

\begin{abstract}
Non-negative blind source separation (BSS) has raised interest in various fields of research, as testified by the wide literature on the topic of non-negative matrix factorization (NMF). In this context, it is fundamental that the sources to be estimated present some diversity in order to be efficiently retrieved. Sparsity is known to enhance such contrast between the sources while producing very robust approaches, especially to noise. In this paper we introduce a new algorithm in order to tackle the blind separation of non-negative sparse sources from noisy measurements. We first show that sparsity and non-negativity constraints have to be carefully applied on the sought-after solution. In fact, improperly constrained solutions are unlikely to be stable and are therefore sub-optimal. The proposed algorithm, named nGMCA (non-negative Generalized Morphological Component Analysis), makes use of proximal calculus techniques to provide properly constrained solutions. The performance of nGMCA compared to other state-of-the-art algorithms is demonstrated by numerical experiments encompassing a wide variety of settings, with negligible parameter tuning. In particular, nGMCA is shown to provide robustness to noise and performs well on synthetic mixtures of real NMR spectra.
\end{abstract}


\section{Introduction}

\IEEEPARstart{I}{n} many applications, such as nuclear magnetic resonance (NMR) spectrometry or mass-spectrometry, measurements are often made of mixtures of physical components which can be identified by their specific spectrum. Discriminating between these elementary components or sources can be made by acquiring several measurements at different times or locations in order to observe different mixtures, yielding multispectral data.

In this context, blind source separation (BSS) aims at recovering the spectra from measurements in which the components sources are mixed up together in an unknown way. The instantaneous linear mixture model assumes that the $m$ measurements $y_{i,\cdot}$ are linear mixture of $r$ sources with spectra $s_{j,\cdot}\in \mathbb{R}^{1 \times n}$. In other words, there exist mixture coefficients $(a_{ij})$ such that:
\begin{equation}
y_{i,\cdot} = \sum_{j=1}^r a_{ij} s_{j,\cdot}+z_{i,\cdot}~,~\forall i\in\{1,.., m \},
\end{equation}
where the vectors $z_{i,\cdot}$ are added in order to account for noise and model imperfections. This mixing model can be conveniently rewritten in matrix form: 
\begin{equation}
\label{eq:BSSeq}
\bY=\bA\bS +\bs{Z} ,
\end{equation}
where:
\begin{itemize}
\item $m$ is the number of measurements.
\item $n$ is the number of samples of the sources.
\item $r$ is the number of sources.
\item $\bY  \in \mathbb{R}^{m \times n}$ is the measurements matrix in which each row is a measurement.
\item $\bS \in \mathbb{R}^{r \times n}$ the unknown source matrix in which each row is a spectrum/source.
\item $\bA \in \mathbb{R}^{m \times r}$ the unknown mixing matrix which defines the contribution of each source to the measurements.
\item $\bs{Z} \in \mathbb{R}^{m \times n}$ is an unknown noise matrix accounting for instrumental noise and/or model imperfections.
\end{itemize}
\medskip

With the notation $\| \bs{X} \|_p = \sqrt[p]{\sum_{ij}|x_{ij}|^p}$ (Frobenius norm for $p=2$) and for independent and identically distributed Gaussian noise, the maximum-likelihood estimate is then provided by the standard problem:
\begin{equation}
\label{eq:BSS}
\underset{\bs{A},~\bs{S}}{\text{argmin}}~\frac{1}{2}\|\bs{Y}-\bs{A} \bs{S}\|_2^2.
\end{equation}

However, this problem presents an infinite number of solutions which are not necessarily of any interest with respect to a given application. It is therefore standard to constrain $\bA$ and/or $\bS$ so as to limit the search to minima with a desired structure. In this article, we focus on the assumption that both $\bA$ and $\bS$ are non-negative, yielding Non-negative Matrix Factorization (NMF).\medskip

The non-negativity assumption arises naturally in many applications, including text mining \cite{Berry_05_EmailSurveillanceUsing}, clustering \cite{Kim_06_SparseNonnegativeMatrix}, audio processing \cite{Fevotte_09_Nonnegativematrixfactorization} and spectrometry \cite{Dubroca_12_WeightedNMFhigh}. Indeed, the sources in $\bS$ can for instance be mass or power spectra, which are non-negative, and the mixtures can represent concentrations, which cannot be negative either. The problem is written under the constrained form:

\begin{equation}
\label{eq:NMF}
\underset{\bA\ge \b0,~\bS\ge \b0}{\text{argmin}}~\mathcal{D}(\bY||\bA\bS).
\end{equation}

$\mathcal{D}(.||.)$ can be the $\ell_2$ distance such as in \eqref{eq:BSS} or any divergence taking into account other priors on the noise \cite{Fevotte_09_Nonnegativematrixfactorization, Lee_99_Learningpartsobjects, Cichocki_06_Csiszarsdivergencesnon}. In this article, we focus on Gaussian noise and therefore use the $\ell_2$ distance.\medskip

The first publications dealing with this type of problems comes from Paatero \& Tapper \cite{Paatero_94_Positivematrixfactorization} and Lee \& Seung \cite{Lee_99_Learningpartsobjects} who provided convergent gradient descent type algorithms. They emphasized on the importance of the non-negativity constraint which yields a ``part-based" representation of the data. Indeed, only additive/constructive, and not subtractive/destructive combinations are then allowed in order to represent the data.\medskip

NMF shares a couple of indeterminacies with other BSS problems:
\begin{itemize}
\item \textit{permutations}: if $\pi$ is a permutation, $\bA\bS=\sum_i a_{\cdot,i} s_{i,\cdot}=\sum_i a_{\cdot,\pi(i)}s_{\pi(i),\cdot}$ so that one cannot hope to recover the order of the rows of $\bS$ and columns of $\bA$.
\item \textit{scales}: if $\bs{\Delta} \in \mathbb{R}^{r\times r}$ is diagonal with strictly positive diagonal coefficients, $\bA\bS=(\bA\bs{\Delta})(\bs{\Delta}^{-1}\bS)$ so that if $(\bA,\bS)$ is a solution of \eqref{eq:NMF}, $(\bA\bs{\Delta},\bs{\Delta}^{-1}\bS)$ is also solution.
\end{itemize}

However, unlike problem \eqref{eq:BSS} for which a solution can be conveniently obtained through SVD decomposition for instance, NMF is NP-Hard \cite{Vavasis_09_ComplexityNonnegativeMatrix} and can present numerous local minima. For this reason,  additional constraints or priors information can be helpful to recover the sought sources. In the next sections, we impose a sparse prior on the sources, together with their non-negativity.

\subsection*{Contributions}
This article details and extends the preliminary work in \cite{Rapin_12_RobustNonNegative} with more complete experimental study and a more advanced understanding of the algorithms. Specifically, noise is accurately accounted for. It illustrates the proposed approach on realistic data as well. In Section~\ref{sec:soa}, we first give a review of the state-of-the-art NMF methods and sparse NMF techniques proposed so far. Our approach aims at obtaining a better management of the noise through an ad hoc regularization strategy. For that purpose, we introduce in Section~\ref{sec:secnGMCA} an extension of the sparse BSS algorithm GMCA (Generalized Morphological Component Analysis) to tackle the problem of sparse and non-negative BSS from noisy data. This extension is motivated by the robustness of GMCA to additive noise contamination.\medskip

 We further show that a special care has to be taken to constrain both the sparsity and the non-negativity of the signals. More specifically, improperly constrained methods can lead to unstable and sub-optimal solutions. We therefore introduce the non-negative GMCA (nGMCA) which has the ability to provide exact solutions of the constrained and penalized sub-problems, thanks to proximal calculus methods. Last but not least, the proposed algorithm is a simple to use technique since it barely requires any parameter tuning.\medskip
 
 In Section~\ref{sec:Experiments} the proposed approach is compared with standard NMF and sparse NMF algorithms. Exhaustive experiments show that nGMCA allows for better robustness to noise, large numbers of sources and the conditioning of the mixing matrix, and therefore outperforms most algorithms for a broad variety of settings. Finally, in section \ref{sec:Application}, we illustrate how nGMCA performs also well to separate out synthetic NMR spectra.

\section{State-of-the-art NMF methods}
\label{sec:soa}

\subsection{Standard NMF algorithms}

The minimization of  the cost function from Problem \eqref{eq:NMF} is generally solved by alternately updating $\bA$ and $\bS$. Indeed, while this problem is not convex, the sub-problem in $\bA$ writes: 
\begin{equation}
\label{eq:subNMF}
\underset{\bA\ge \b0}{\text{argmin}}~\mathcal{D}(\bY||\bA\bS)
\end{equation}
and the equivalent sub-problem in $\bS$ are often convex and their minimization is therefore much easier. These alternating updates are then repeated for a large number of iterations.\medskip

The first converging NMF algorithm, designed by Lee \& Seung \cite{Lee_99_Learningpartsobjects,Lee_01_Algorithmsnonnegative}, updates $\bA$ and $\bS$ with a weighted gradient descent. The weights insure that the gradient steps do not increase the cost function in \eqref{eq:NMF} and keep $\bA$ and $\bS$ non-negative, since the update can be recast as a pointwise product of non-negative matrices. With $\odot$ the element-wise matrix multiplication and $\oslash$ the element-wise matrix division, the update rule for $\bA$ and $\bS$ in the case of a least square cost function can be written as follows:
\begin{align}
        \bA_{k+1}&\leftarrow \bA_{k}\odot(\bY\bS_{k}^T)\oslash(\bA_{k}\bS_{k}\bS_{k}^T),\notag \\
        \bS_{k+1}&\leftarrow \bS_{k}\odot(\bA_{k+1}^T\bY)\oslash(\bA_{k+1}^T\bA_{k+1}\bS_{k}).\label{eq:multiplicative_update_rule}
\end{align}

The multiplicative update rule is usually considered as a standard because of its convenience, with no parameter to set. As well, it was the first convergent algorithm proposed to solve the NMF problem. However, it has been shown to be slow and that the monotone decrease of the algorithm does not insure convergence to a local minimum \cite{Gonzalez_05_AcceleratingLeeSeung,Berry_07_Algorithmsandapplications}. First-order methods are also used in projected/proximal gradient descent algorithms \cite{Xu_11_Alternatingproximalgradient}, interior point gradient \cite{Merritt_05_InteriorPointGradient}, and quasi-Newton algorithms \cite{Zdunek_07_Nonnegativematrixfactorization}.\medskip

Another standard approach is the Alternating Least Square (ALS) algorithm explained in \cite{Paatero_94_Positivematrixfactorization}, which solves exactly the unconstrained cost function with a pseudo-inverse and projects the result on the non-negativity constraint:
\begin{align}
        \bA_{k+1}&\leftarrow \big[\bY \bS_{k}^T(\bS_{k}\bS_{k}^T)^{-1}\big]_+,\notag \\
    	\bS_{k+1}&\leftarrow \big[ (\bA_{k+1}^T\bA_{k+1})^{-1}\bA_{k+1}^T\bY\big]_+,\label{eq:ALS_update_rule}
\end{align}
with the operator $[x]_+= \text{max}(x,0)$. This algorithm is also widely used due to its easy implementation and its efficiency in decreasing the cost function. However, it does not necessarily converge. In Hierarchical ALS (HALS, \cite{Cichocki_07_HierarchicalALSAlgorithms,Cichocki_09_NonnegativeMatrixand}), the columns of $\bA$ and rows of $\bS$ are processed one by one. This yields a simple and fast optimization process to solve the constrained sub-problems.\medskip 

It is however possible to solve exactly the constrained sub-problems of type \eqref{eq:subNMF} at each iteration as follows:
\begin{align}
        \bA_{k+1}&\leftarrow \underset{\bA\ge \b0}{\text{argmin}}~\mathcal{D}(\bY||\bA\bS_k),\notag\\
    	\bS_{k+1}&\leftarrow \underset{\bS\ge \b0}{\text{argmin}}~\mathcal{D}(\bY||\bA_{k+1}\bS)\label{eq:exact_update_rule}.
\end{align}
In \cite{Lin_07_Projectedgradientmethods} for instance, Lin uses a projected gradient descent subroutine to solve the sub-problems. Guan et al.\@ later provided a faster first-order method \cite{Guan_12_NeNMFOptimalGradient}.\medskip

It has to be mentioned that other approaches based on geometrical methods have also been investigated to solve NMF problems\cite{Babaie-Zadeh_04_GeometricApproachSeparating, Ouedraogo_12_Geometricalmethodusing,Gillis_12_FastandRobust}; these approaches are however generally very sensitive to noise.\medskip

As was previously stated, non-negativity is not always sufficient to recover the actual sources and mixing matrix. In non-negative ICA \cite{Plumbley_04_NonnegativePCA}, one additionally enforces the independence of the sources. However, this approach is also sensitive to noise. Sparsity, on the other hand, has been shown to provide robustness to noise. We give a short introduction about this prior in the next section, before presenting sparse NMF algorithms.

\subsection{Sparsity \& NMF}
\subsubsection{An Introduction to Sparsity}~\\
\label{sec:sparsity}

Sparsity constraints have already proved their efficiency to solve a very wide range of inverse problems (see \cite{Starck_10_SparseImageand} and references therein).  In the context of BSS, sparsity has been shown to increase the diversity between the sources which greatly helps their separation \cite{Zibulevsky_99_BlindSourceSeparation, Li_03_Sparserepresentationand,Bobin_07_Sparsityandmorphological}. In the wide sense, a sparse signal is such that its information content is concentrated into only a few large non-zero coefficients, or can be well approximated in such a way. The sparsity of a signal however depends on the basis or dictionary in which it is expressed. For instance, a sine wave will be sparse in the Fourier domain since it can be encoded with one coefficient in this domain, while in the direct domain most of its coefficients are non-zero: the more a basis captures the structures of a signal, the sparser will be in such a basis. In this article we will however only focus on sparsity in the direct domain.\medskip

\subsubsection{Sparse NMF Algorithms}~\\
\label{sec:sparseAlgorithms}
In many applications the non-negativity and sparsity of the sources arise naturally as for instance in MS or NMR spectroscopy. Recent works have emphasized the fact that this knowledge can indeed help perform more relevant factorizations \cite{Kim_06_SparseNonnegativeMatrix, Eggert_04_Sparsecodingand, Hoyer_02_Nonnegativesparse}.\medskip

In \cite{Kim_06_SparseNonnegativeMatrix}, Kim \& Park have proposed to formulate the NMF problem as:
\begin{equation}
\label{eq:KimNMF}
\underset{\bA\ge \b0,\bS\ge \b0}{\text{argmin}}~\|\bY-\bA\bS\|_2^2+\eta \|\bA\|_2^2+\beta \sum_{t=1}^n \|s_{\cdot ,t}\|_1^2.
\end{equation}
In this equation, the sparsity-enforcing regularizer $\sum_{t=1}^n \|s_{\cdot ,t}\|_1^2$ favors solutions where a single source dominates at each sample. However, it does not enforce the intrinsic sparsity of each of the sources. The authors made use of an active-set method to solve this problem. This technique can solve exactly each constrained sub-problem in a similar way than in Problem \eqref{eq:exact_update_rule}. However, it is not clear how the parameters $\eta$ and $\beta$ must be set. The authors provide an implementation on their website\footnote{\url{http://www.cc.gatech.edu/~hpark/nmfsoftware.php}}.   

In \cite{Zdunek_07_Nonnegativematrixfactorization}, Zdunek \& Cichocki used a similar regularization term for $\bS$ but with  $\beta$ decreasing during the algorithms in order to be more robust to local minima, and without exactly solving the sub-problems.\medskip

In \cite{Hoyer_02_Nonnegativesparse}, Hoyer used a sparse regularization of the form $\lambda \|\bS\|_1$ that uniformly enforce the sparsity of $\bS$.  The author later used a different type of sparse prior in \cite{Hoyer_04_Nonnegativematrix} defined for some vector $x \in \mathbb{R}^n$ as follows:
\begin{equation}
\label{eq:HoyerSparseness}
\text{sparseness}(x)=\frac{\sqrt{n}-\frac{\|x\|_1}{\|x\|_2}}{\sqrt{n}-1}.
\end{equation}
This sparseness function goes from 1 when $x$ is perfectly sparse ---only 1 active coefficient--- to 0 when it is not at all ---all coefficients active, with the same value. The idea is therefore to optionally impose a chosen level of sparsity for the sources and/or the mixtures:
\begin{equation}
\left\{
    \begin{array}{ll}
        \underset{\bA\ge \b0,\bS\ge \b0}{\text{argmin}}~\|\bY-\bA\bS\|_2^2,\\ 
    		\text{sparseness}(a_{\cdot,i})=\lambda_{A},~\forall i\text{~(optional)},\\
    		\text{sparseness}(s_{i,\cdot})=\lambda_{S},~\forall i\text{~(optional)}.\label{eq:HoyerNMF}
    \end{array}
\right.
\end{equation}
This problem is solved by using projected gradient descent steps followed by a projection on the sparseness constraint if the constraint is active and with a multiplicative update otherwise. However, the constraint is directly related to the expected sparseness level of the sources which is not necessarily known beforehand. Furthermore, hard-constraining the sparseness level may make the solution very dependent on the sparseness parameters $\lambda_A$ and $\lambda_S$. An implementation of this algorithm is available online\footnote{\url{http://www.cs.helsinki.fi/u/phoyer/software.html}}.\medskip

Sparse HALS \cite{Cichocki_07_HierarchicalALSAlgorithms,Cichocki_09_NonnegativeMatrixand} aims at solving Problem \eqref{eq:NMF} with a sparsity penalization of the form $\lambda \|\bS\|_1$. In the case of an $\ell_2$ data fidelity term, this problem can be rewritten as: 
\begin{equation}
\label{eq:HALS}
\underset{a_{\cdot,i}\ge 0,~s_{i,\cdot}\ge 0}{\text{argmin}}~\|\bY-\sum_{i=1}^r a_{\cdot,i}~s_{i,\cdot}\|_2^2+ \lambda \sum_{i=1}^r \big( \sum_{j=1}^n  s_{i,j} \big).
\end{equation}
In this algorithm, the columns of $\bA$ and lines of $\bS$ are updated one by one. As each sub-problem admits a straightforward analytic solution and no matrix inversion is required, the HALS algorithm turns to be a simple and fast NMF solver. In a recent implementation of the HALS \cite{Gillis_12_AcceleratedMultiplicativeUpdates, Gillis_12_SparseandUnique}\footnote{\url{https://sites.google.com/site/nicolasgillis/code}}, the parameter $\lambda$ of  is automatically managed in order to obtain a required sparsity rate (defined as the ratio of coefficients smaller than $10^{-6}$ times the largest coefficient).
\medskip

It is important to note that none of these algorithms makes use of the sparse prior explicitly to deal with additive noise; therefore they may not be robust in case of noise contamination.

\section{Non-negative Generalized Morphological Component Analysis}
\label{sec:secnGMCA}

\subsection{A first naive extension}
\label{sec:nGMCA}

In the last decade the use of sparsity in the field of BSS has been widely explored. In \cite{Bobin_07_Sparsityandmorphological,Bobin_08_BlindSourceSeparation}, the authors have introduced a sparsity-enforcing  BSS technique coined Generalized Morphological Component Analysis (GMCA) which has been shown to be effective at separating out sparse signals from noisy data. Morphological diversity has been defined in \cite{Bobin_08_BlindSourceSeparation} as a mean to characterize separable sources based on their geometrical structures or morphologies : separable sources with different morphologies do not share the same significant coefficients in a given sparse representation. When sparsity holds in the direct domain this means that the entries of each source with the most significant amplitudes should be different. This does not mean that their supports are disjoint but rather their most significant elements should be disjoint.

The objective of this paper is to extend this sparsity-enforcing BSS algorithm to deal with non-negative mixtures. Following \cite{Bobin_07_Sparsityandmorphological}, the GMCA with an additional non-negative constraint estimates the mixing matrix and the sources by minimizing the following optimization problem:

\begin{equation}
\label{eq:GMCA}
\underset{\bA\ge \b0,~\bS\ge \b0}{\text{argmin}}~\frac{1}{2}\|\bY-\bA \bS\|_2^2+\lambda \|\bS\|_0.
\end{equation}
The $\ell_0$ pseudo-norm counts non-null coefficients in $\bS$ and therefore limits their number, thus enforcing the sparsity of $\bS$. In the vein of Alternating Least Squares, GMCA alternately and iteratively estimates the unconstrained least square solution and projects on the non-negativity constraint, with an additional thresholding step for the sources in order to keep only the most significant coefficients. These updates are provided in lines 6 and 7 of \textbf{Algorithm \ref{alg:nGMCAnaive}} where the hard-thresholding operator $\text{Hard}_{\lambda}$ is defined as follows:
\begin{equation}
\label{eq:hardthresholding}
\text{Hard}_\lambda: x \mapsto \begin{cases}
        0 \text{ if } |x|<\lambda , \\
        x \text{ otherwise}.
    \end{cases}
\end{equation}

It has been emphasized in \cite{Bobin_07_Sparsityandmorphological} that one crucial feature of GMCA is the use of a decreasing threshold $\lambda$. At the beginning, this parameter is first set to a high value and then decreases throughout the iterative down to a final value that depends on the noise level. Simulated annealing has already inspired decreasing $\ell_2$ and $\ell_{1,2}$ regularizations in NMF \cite{Zdunek_07_Nonnegativematrixfactorization, Cichocki_07_RegularizedAlternatingLeast}. The motivation behind a decreasing threshold is however different:
\begin{enumerate}
\item first estimating the mixing matrix from the entries of the sources that have the highest amplitude and thus likely to belong to only one source.
\item help removing the smallest coefficients which are more sensitive to noise contamination.
\end{enumerate}
\smallskip
In the same way as in the original GMCA, for a source with index $i$, the threshold $\lambda_i$ is set to $\tau_\sigma  \sigma_i^\text{source}$ where $\sigma^\text{source}_i$ is an empirical estimator (the median absolute deviation) of the source noise variation. $\tau_\sigma$ is chosen at each iteration in order to obtain a linear increase of the number of active coefficients in $\bS$, so as to refine the estimation while maintaining some continuity. The final $\tau_\sigma$, $\tau_\sigma^\infty$, is usually taken in the range $[1,3]$ as a trade-off between sufficient denoising and correct separation. Indeed, for a sparse signal contaminated by i.i.d.\@ Gaussian noise with standard deviation $\sigma$, thresholding at 3$\sigma$ rejects noise samples with probability 0.99. Still, in BSS, too large a final threshold could leave some leakage between the sources. This nGMCA will be considered as a {\it naive} extension of GMCA and used as a reference algorithm. It will be coined naive non-negative GMCA (nGMCA$^\text{naive}$).\medskip

\begin{algorithm}[!t]
\caption{: nGMCA$^\text{naive}$}
\label{alg:nGMCAnaive}
\begin{algorithmic}[1]
\Require $\bY$, $K$
\State \textbf{initialize} $\bA$ and $\bS$
\For{$k\leftarrow 1,K$}
\State Normalize the columns of $\bA_{k-1}$
\State $\bS_\text{all}=(\bA_{k-1}^T\bA_{k-1})^{-1}\bA_{k-1}^T\bY$
\State Select the thresholds $\lambda_{k}$ considering $\bS_\text{all}$
\State $\bS_{k}\gets\big[\text{Hard}_{\lambda_{k}}(\bS_\text{all})\big]_+$
\State $\bA_{k}\gets \big[\bY \bS_{k}^T(\bS_{k}\bS_{k}^T)^{-1}\big]_+$
\EndFor
\State \textbf{return} $\bA_K,~\bS_K$
\end{algorithmic}
\end{algorithm}

\paragraph*{Limitations} It is very important to notice that this algorithm, as based on alternating projected least-squares, is only a proxy which generally does not converge to a stable couple $(\bA,\bS)$, hence the ``naive" denomination. This means that the solution given by this type of algorithm may not be stable and be sub-optimal. The solution may not provide the sparsest non-negative sources. This is due to the fact that it deals with the data fidelity term and the constraints in a completely independent way, thus not exactly solving the sub-problems such as in Problem \eqref{eq:exact_update_rule}.\medskip

Next, we therefore propose an alternative algorithm which exactly solves the non-negatively constrained and $\ell_1$ penalized sub-problems and should allow for more robust and stable solutions.

\subsection{nGMCA}
\label{sec:RnGMCA}

We first propose to tackle the sparse non-negative BSS problem using an $\ell_p$ sparse regularization with $p\in \{0,1\}$ such as in \eqref{eq:GMCA}. This then amounts to solving the following optimization problem:
\begin{equation}
\label{eq:RnGMCA}
\underset{\bA,~\bS}{\text{argmin}}~\frac{1}{2}\|\bY-\bA \bS\|_2^2+\lambda \|\bS\|_p+i^+(\bS)+i^+(\bA),
\end{equation}
where $i^+$ is the characteristic function of the non-negative orthant, which enforces the non-negative constraints. The characteristic function of the non-negative orthant is defined as:
\begin{align}
i^+: x \mapsto &\begin{cases}
        0 \text{ if } x \ge 0,\\
        +\infty \text{ otherwise}.
    \end{cases}
\end{align}
In the previous section, we emphasized that a naive approach based on projected least-square does not necessarily provide a stable and thus optimal solution to the problem. To go beyond the aforementioned naive extension of GMCA, one has to alternatively and exactly minimize the constrained sub-problems in $\bA$ and $\bS$ so as to obtain stable solutions with the sought structure. Let's first have a look at the sub-problem in $\bS$; assuming $\bA$ is fixed, the sources are estimated as follows:
\begin{equation}
\label{eq:subps}
\underset{\bS}{\text{argmin}}~\frac{1}{2}\|\bY-\bA \bS\|_2^2+\lambda \|\bS\|_p+i^+(\bS).
\end{equation}

This problem minimizes a function which can be split into the sum of a differentiable quadratic term and two non-smooth and non-differentiable terms: the $\ell_p$ norm and the characteristic function of the non-negative orthant $i^+$. The choice $p=1$ provides the closest convex surrogate to the sparse $\ell_0$ norm and in this case, the problem admits a unique solution which, however, cannot be formulated explicitly. Fortunately, thanks to recent advances in proximal calculus and splitting techniques, efficient algorithms can be designed to solve this type of problems.\medskip

\subsubsection{Proximal calculus and splitting methods}
\label{sec:proximal}

Proximal splitting methods \cite{Combettes_05_Signalrecoveryby} aim at minimizing convex functions which may not be differentiable. The idea of these methods is to split the cost function into the sum of several convex functions which are alternately locally minimized. Many problems can be decomposed into the general form:
\begin{equation}
\label{eq:FBequation}
\underset{\bS}{\text{argmin}}~f(\bS)+g(\bS),
\end{equation}
where $f(\bS)$ is a smooth and convex data fidelity term; and $g(\bS)$ a convex regularization term which may not be differentiable.\medskip

Since $f$ is differentiable, it can be locally minimized with a gradient descent step. When $g$ is not, but if it is convex, proper and lower semi-continuous, one can define the proximal operator of $g$:
\begin{equation}
\label{eq:proximal}
\text{prox}_g:x\mapsto \underset{y}{\text{argmin}}~\frac{1}{2}\|y-x\|_2^2+g(y).
\end{equation}
In this case, the following process called forward-backward splitting algorithm (FBS, see \cite{Combettes_05_Signalrecoveryby}):
\begin{equation}
\label{eq:FBupdate}
x_{k+1}=\text{prox}_{\frac{1}{L}g}(x_k-\frac{1}{L}\nabla f(x_k))
\end{equation}
has been shown to converge to the solution of Problem \eqref{eq:FBequation} if $\nabla f$ is $L$-Lipschitz.\medskip

The update of $\bA$ is made by solving the following problem:
\begin{equation}
\label{eq:subpa}
\underset{\bA}{\text{argmin}}~\frac{1}{2}\|\bY-\bA \bS\|_2^2 + i^+(\bA),
\end{equation}
which can be recast in the general form \eqref{eq:FBequation} by defining $f({\bA}) = \frac{1}{2}\|\bY-\bA\bS\|_2^2$ and $g({\bA}) = i^+(\bA)$. In this case, the gradient of $f$ is trivially equal to: $\nabla f({\bA}) = (\bA\bS- \bY){\bS}^T$, it is $L$-Lipschitz with $L = \| {\bS \bS^T} \|_{s}$ where $\|~.~\|_s$ is the matrix spectral norm. The proximal operator of $g$ is the projector onto the non-negative orthant $[~.~]_+$. The FBS then reduces to a projected gradient algorithm similar to the updates in \cite{Lin_07_Projectedgradientmethods}.

Similarly, the update of $\bS$ in Problem~\eqref{eq:subps} with $p=1$ can be solved with FBS, with $g({\bS}) = \lambda \|\bS\|_1 +i^+(\bS)$. The proximal operator of this function also takes an explicit form usually termed  skewed-position soft-thresholding operator:

\begin{equation}
\label{eq:positiveSoftThresholding}
\text{prox}_{\lambda \|~.~\|_1+i^+(~.~)}:x\mapsto \big[\text{Soft}_\lambda (x)\big]_+,
\end{equation}
where the soft-thresholding operator is defined as:
\begin{equation}
\text{Soft}_\lambda: x \mapsto \text{sign}(x)[|x|-\lambda]_+.
\end{equation}

This soft-thresholding, induced by the $\ell_1$ norm, is well-known to introduce a bias. It is therefore customary to use $p=0$. In the FBS this is made by replacing the soft-thresholding operator with the hard-thresholding. Rigorously, it is not a proximal operator since $\|~.~\|_0$ is not convex nor semi-continuous; this means that there is no convergence guarantee of the forward-backward splitting algorithm when hard-thresholding is used.\medskip

\subsubsection{Description of nGMCA}

In the same vein as the naive nGMCA introduced in the previous section, $\bA$ and $\bS$ are alternately updated with the exception that, now, each sub-problem is solved exactly, which guarantees that the solution $(\bA, \bS)$ is stable and has the sought structure. The main steps of the algorithm are given in \textbf{Algorithm \ref{alg:nGMCAS}}. One must notice that exactly solving each sub-problem is costly since it requires sub-iterations at each step. Fortunately, it has been recently showed that the speed of convergence of the FBS algorithm can be greatly improved by using the multi-step techniques introduced by Nesterov \cite{Nesterov_12_Gradientmethodsminimizing}. Both steps have the algorithm make use of an accelerated version of the FBS algorithm \cite{Beck_09_FastIterativeShrinkage}. A detailed description of how this accelerated algorithm is used to update $\bS$ is given in appendix \ref{app:FISTAsubproblem}.\medskip

A special care has to be given to renormalizations since the $\ell_1$ regularization tends to reduce the norm of $\bS$. More specifically, since the norm of $\bA$ can keep increasing to compensate the reduction of the norm of $\bS$, the algorithm can converge to the degenerate solution $\bS=\b0$ and $\bA=\bs{\infty}$. In the algorithm, the columns of $\bA$ are therefore renormalized to $\ell_2$ unity before updating $\bS$. This also assigns to the coefficients of $\bS$ their overall importance in the estimation of $\bY$.\medskip

Following the general thresholding strategy used in GMCA and its extensions, the threshold $\lambda$ decreases from step to step. However, the strategy used in this version nGMCA differs from the one used in the naive approach. In the former, the threshold is applied to the sources as defined by their least-square estimate. On the contrary the threshold in nGMCA$^\text{S}$ applies at each gradient descent step. The update rule of $\bS$ in the sub-iterations of nGMCA$^\text{S}$ (without the acceleration) can be written as follows:
\begin{equation}
\label{eq:softSubiteration}
\bS_{k+1}\leftarrow \big[\bS_k-\frac{1}{L}\big(~\bA^T(\bA\bS_k-\bY)-\lambda \mathds{1}_{rn}\big)\big]_+,
\end{equation}
with $\mathds{1}_{rn}\in \mathbb{R}^{r \times n}$ containing only ones. Iterative soft-thresholding therefore operates on the gradient and not directly on the source values like in the naive nGMCA. Also, unlike with hard-thresholding, a variation of $\lambda$ affects all active coefficients. Our strategy consists in starting with a large parameter ${\lambda_0=\|\bA_0^T(\bA_0\bS_{0}-\bY)\|_\infty}$ which forces the coefficients of $\bS$ to be non-increasing in the first iteration. The threshold is then linearly decreased in order to refine the solution while preserving continuity, down to $\tau_\sigma^\infty \sigma^\text{grad}_i$ where $\sigma^\text{grad}_i$ is this time an estimate of the noise level in the $i^\text{th}$ row of the gradient.\medskip

We also implemented a version of nGMCA using hard-thresholding aiming at solving Problem \eqref{eq:RnGMCA} with an $\ell_0$ pseudo-norm instead of the $\ell_1$ norm for the regularization. The superscript $^\text{H}$ and $^\text{S}$ are specified to differentiate between respectively the hard- and the soft-thresholding versions.

\begin{algorithm}[!t]
\caption{: nGMCA$^\text{S}$}
\label{alg:nGMCAS}
\begin{algorithmic}[1]
\Require $\bY$, $K$
\State \textbf{initialize} $\bA_0$, $\bS_0$ and $\lambda_1$
\For{$k\leftarrow 1,K$}
\State Normalize the columns of $\bA_{k-1}$
\State $\bS_{k}\gets \underset{\bS\ge \b0}{\text{argmin}}~\frac{1}{2}\|\bY-\bA_{k-1}\bS\|_2^2+\lambda_{k} \|\bS\|_1$
\State $\bA_{k}\gets \underset{\bA\ge \b0}{\text{argmin}}~\frac{1}{2}\|\bY-\bA\bS_{k}\|_2^2$ 
\State Select $\lambda_{k+1}\le\lambda_k$
\EndFor
\State \textbf{return} $\bA_K,~\bS_K$
\end{algorithmic}
\end{algorithm}

\section{Numerical experiments}
\label{sec:Experiments}

In this section we first compare the introduced algorithms and classical algorithms on noiseless data, in order to better understand their behaviors, and then we benchmark the GMCA-based algorithms with state-of-the-art sparse algorithms on noisy data. The settings of the simulations and the evaluation methodology are described in the following sections.

\subsection{Settings}

Reference matrices $\bA^\text{ref} $ and $\bS^\text{ref}$ coefficients are uniformly generated respectively from the distribution of $|B_{p_A}G_{\alpha_A}|$ and $|B_{p_S}G_{\alpha_S}|$, where:
\begin{itemize}
\item $B_p$ is a Bernoulli random variable with activation parameter $p$, i.e. it equals 1 with probability $p$ and 0 otherwise.
\item $G_\alpha$ is a centered and reduced Generalized Gaussian random variable with shape parameter $\alpha$. 
\end{itemize}
In practice $p$ and $\alpha$ control 2 kinds of sparsity. The Bernoulli parameter affects the number of actual zeros in $\bA$ and $\bS$ and therefore exact sparsity. On the other hand, $\alpha$ selects the sharpness of the distribution of $G_\alpha$, which pdf is proportional to $e^{\frac{-|x-\mu|^\alpha}{\beta}}$ (with $\mu=0$ and $\beta$ dependent on the standard deviation which is fixed to 1 here). As special cases, for $\alpha=2$, $G_\alpha$ is a Gaussian random variable, and with $\alpha=1$ it is a Laplacian random variable. With $\alpha \le 1$, $G_\alpha$ is considered as approximately sparse ---the generated signals become sparser when $\alpha$ decreases.

In the experiments and unless stated otherwise, the standard settings will be $p_A=1$, $\alpha_A=2$, $\alpha_S=1$ with $m=200$ measurements of $n=200$ samples.

\subsection{Evaluation of the Results}

In order to evaluate and compare the algorithms, a scale and permutation invariant criterion is needed. This criterion has to be well adapted to measure the reconstruction performance. In many applications, the signal of interests are the sources. More precisely, it has to be noticed that noise-reducing priors are applied on the sources only. This implies that a good estimate of the sources should be the least contaminated by noise and interferences from the other sources. Moreover, in a noisy setting, a perfectly estimated mixing matrix $\bA$ do not necessarily yield a good estimate of the sources: indeed a slightly degraded mixing matrix may be preferred if it leads to less noisy sources. These points make criteria based on this variable not adequate to measure a good separation. In the next, we will focus on estimating the separation performance using a criterion on the sources $\bS$.\medskip

In \cite{Vincent_06_Performancemeasurementin}, Vincent et al.\@ have proposed different criteria to evaluate the performance of blind source separation techniques. In the noisy setting, they propose separating each estimated source $s^\text{est}$ into the sum of several components:
\begin{equation*}
s^\text{est}=s_\text{target}+s_\text{interf}+s_\text{noise}+s_\text{artifacts},
\end{equation*}
with $s_\text{target}$ the projection of $s^\text{est}$ on the reference source it estimates; and $s_\text{interf}+s_\text{noise}+s_\text{artifacts}$, in its orthogonal space, respectively standing for interferences with other sources, contamination with noise and contamination with algorithm artifacts.
They design an SNR-type energy ratio criterion named Source Distortion Ratio (SDR):
\begin{equation}
\text{SDR}(s^\text{est})=10~\text{log}_{10}\left(\frac{\|s_\text{target}\|_2^2}{\|s_\text{interf}+s_\text{noise}+s_\text{artifacts}\|_2^2}\right).\label{eq:SDR}
\end{equation}
As stated in \cite{Vincent_06_Performancemeasurementin}, this criterion is a global performance measure taking into account all the elements of the reconstruction, i.e. a correct separation (low $s_\text{interf}$), efficient denoising (low $s_\text{noise}$) and little artifacts left by the algorithm (low $s_\text{artifacts}$). Also, this criterion has the advantage of being scale-invariant. In the next, the SDR will be used to evaluate the separation performance of the proposed technique with respect to state-of-the-art methods.

Because of the permutation invariance, one cannot know \textit{a priori} which estimated source stands for which reference source. Reference and estimated sources are therefore paired one-to-one in order to obtain the best mean SDR. The SDR on $\bS$ (coined SDR$_S$) has be used in the experiments below in order to assess the performances of the algorithms. In the following, the SDR has been evaluated from several Monte-Carlo simulations; the number of simulations will be given in each figure's caption.

\subsection{Behavioral Study: Noiseless Data}
\label{sec:ExpNoiseless}

In this first experimental section, nGMCA$^\text{naive}$, nGMCA$^\text{H}$ and nGMCA$^\text{S}$ are tested on data with a large activation rate for $\bS$ ($p_S=80\%$)  and therefore not the very sparse sources for which they are designed. These settings are not favorable for the nGMCA algorithms, which allows to emphasize the differences of behavior between them. Since the problem reduces to an exact factorization when there is no noise, $\tau_\sigma^\infty$ is set to 0, hence the final threshold in this case is identical for all the algorithms. ALS \cite{Paatero_94_Positivematrixfactorization}, the Multiplicative Update \cite{Lee_01_Algorithmsnonnegative} and (non-sparse) accelerated HALS \cite{Gillis_12_AcceleratedMultiplicativeUpdates} are also performed as standard algorithms to play the role of references. The maximum number of iterations were set to large enough numbers in order to assure the convergence of all the algorithms. Precisely, the number of iterations is set to 5000 for HALS, 40,000 for the Multiplicative Update, 500 for ALS and the GMCA-based algorithms, with a maximum of 80 sub-iterations in nGMCA$^\text{S}$ and nGMCA$^\text{H}$).\medskip

\medskip
\subsubsection{Summary of the experiments}
\begin{itemize}

\item Figure \ref{fig:r_a80n00}:  this  benchmark shows the influence of the number of sources on the estimation of $\bS$. This parameter is of course important since the more numerous they are, the more difficult they are to separate. nGMCA$^\text{naive}$ outperforms ALS, while with $\tau_\sigma^\infty=0$, the final iterations are identical for both algorithms. The thresholding strategy of the first stage of the iterations therefore proves its efficiency for the separation. Though nGMCA$^\text{S}$ is not performing as well as nGMCA$^\text{H}$ and nGMCA$^\text{naive}$ with few sources, it is much more robust than all the algorithms with large numbers of sources. This is further detailed in paragraph \ref{sec:L1vsL0}.
\smallskip

\begin{figure}[!t]
\centering
\includegraphics[width=3.35in]{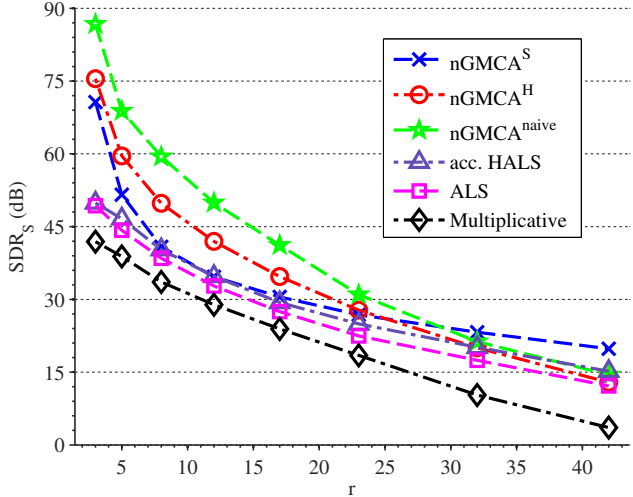}
\caption{Reconstruction SDR (SDR$_S$) with respect to the number of sources $r$  ($p_S=80\%$, noiseless, average of 48 simulations)}
\label{fig:r_a80n00}
\end{figure}

\item Figure \ref{fig:oscillations}: this figure exhibits the evolution of the cost function $\|\bY-\bA\bS\|_2^2$ throughout the iterations, for 40 sources and a large activation rate ($p_S=80\%$). In the refinement phase, the sparsity parameter $\lambda$ is left constant at its final value $\lambda^\infty = \tau_\sigma^\infty \sigma = 0$ in order to observe the convergence of the algorithms and the possibility to enhance the reconstruction. nGMCA$^\text{S}$ converges to a lower value than nGMCA$^\text{H}$ while nGMCA$^\text{naive}$ does not converge at all. An explanation is provided in paragraph \ref{sec:ConstrVsUnconstr}.
\smallskip

\begin{figure}[!t]
\centering
\includegraphics[width=3.35in]{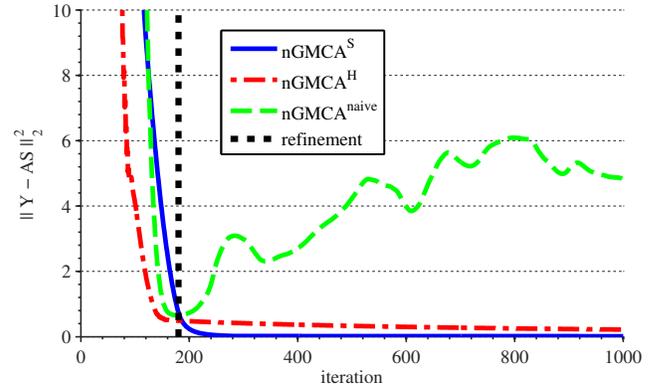}
\caption{Evolution of the cost function $\|\bY-\bA\bS\|_2^2$ during the iterations for a representative example ($p_S=80\%$, $r=40$, noiseless).}
\label{fig:oscillations}
\end{figure}

\item Figure \ref{fig:Aalpha_r35a80}: this benchmark shows the influence of $\bA^\text{ref}$ coefficients distribution on the reconstruction. Modifying the parameter $\alpha_A$ is a way to make its distribution more or less sparse. A sparser $\bA^\text{ref}$ yields less correlated columns and hence a better conditioning of the problem (table \ref{tab:conditioning}) which simplifies the separation. nGMCA$^\text{S}$ tends to be more robust than the other algorithms to ill-conditioned mixtures.

\begin{figure}[!t]
\centering
\includegraphics[width=3.35in]{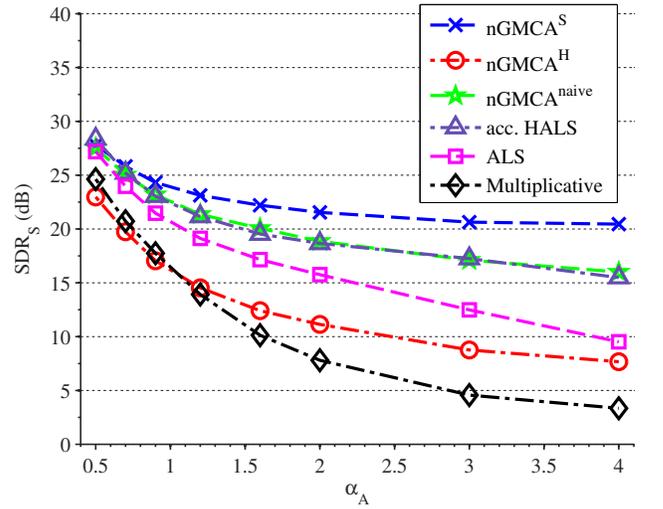}
\caption{Reconstruction SDR (SDR$_S$) with respect to the distribution parameter $\alpha_A$ ($r=35$, $p_S=80\%$, noiseless,  average of 48 simulations)}
\label{fig:Aalpha_r35a80}
\end{figure}

\begin{figure}[!t]
\label{tab:conditioning}
\footnotesize
\begin{tabular}{|l|c|c|c|c|c|c|c|c|}
\hline
 $\boldsymbol\alpha_A$ & 0.50 &0.70 &0.90 &1.20 &1.60 &2.00 &3.00 &4.00\\
\hline
 \bf cond($\bA$) & 6.90 & 8.36 & 9.43 & 10.7 &12.0 & 12.9 & 14.3 &15.3\\
\hline
\end{tabular}
\caption{Conditioning of $\bA^\text{ref}$ with respect to $\alpha_A$ \protect\\($r=35$, average of 48 simulations)}
\end{figure}

\end{itemize}

\medskip
\subsubsection{Properly accounting for the constraints}~\\
\label{sec:ConstrVsUnconstr}
The differences of performance between nGMCA$^\text{naive}$ and nGMCA$^\text{S}$ for large numbers of sources and large activation rates in figure \ref{fig:r_a80n00} can be understood by observing the evolution of the cost function during the iterations (figure \ref{fig:oscillations}). Properly applied non-negativity and sparsity constraints can help refining the reconstruction of $\bA$ and $\bS$ once the sources have been sufficiently disambiguated. Indeed, since nGMCA$^\text{naive}$ does not exactly solve the constrained cost function, it does not necessarily neither converge to a minimum  nor lead to a stable solution, while nGMCA$^\text{S}$ does.

\medskip
\subsubsection{$\ell_1$ Vs $\ell_0$}
\begin{figure}[!t]
\centering
\includegraphics[width=3in]{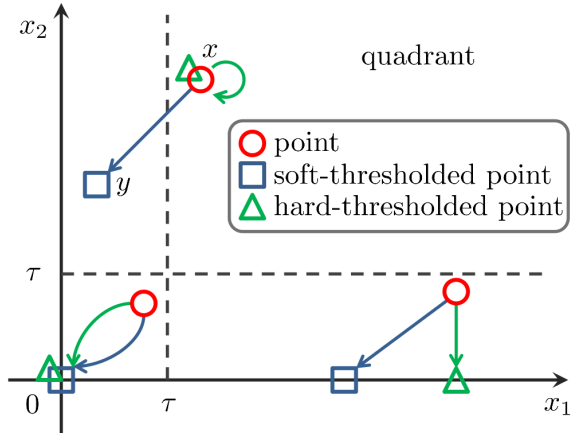}
\caption{Difference between soft- and hard-thresholding (thresholding at $\tau$)}
\label{fig:soft-hard}
\end{figure}

~\\
\label{sec:L1vsL0}
The explanation of the differences between nGMCA$^\text{S}$ and nGMCA$^\text{H}$ lies in the properties of hard- and soft-thresholding. This is summarized in figure \ref{fig:soft-hard} which shows thresholding applied to three two-dimensional points $x=(x_1,x_2)$.  When a point  ---a column of $\bS$--- has two large coefficients, i.e. when the point is in the quadrant, it suffers a bias with soft-thresholding while it remains untouched with hard-thresholding. On the other hand, the shift induced by soft-thresholding increases the ratio of its larger coefficient over its smaller one, which helps reinforce the affectation of a point to the direction of its larger coefficient (in this case: $x_2$). With $y=(y_1,y_2)$ the thresholded point, this means here that $\frac{y_2}{y_1}>\frac{x_2}{x_1}$. While the lesser bias created by hard-thresholding can lead to better accuracy, reinforcing the affectation of each point to a direction with soft-thresholding leads to better separation of the sources.

\begin{figure}[!t]
\centering
\includegraphics[width=3.35in]{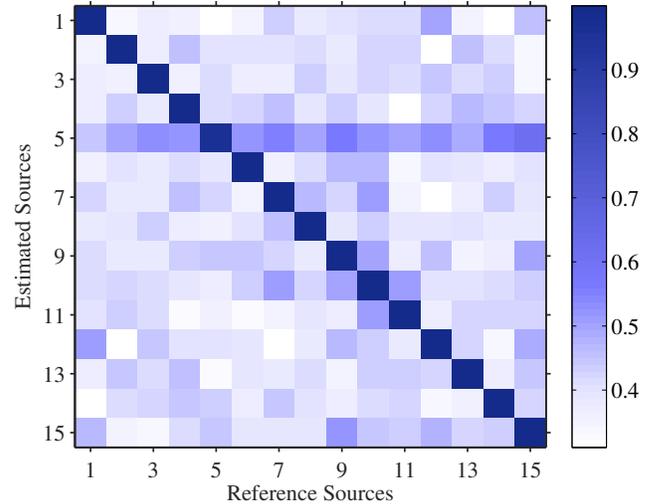}
\caption{Correlation matrix between the estimated and reference sources, in a representative example\protect\\($S^{\text{est}}S^{\text{ref}T}$ after normalization, nGMCA$^\text{S}$, $r=15$, $p_S=80\%$, noiseless)}
\label{fig:correlationBias}
\end{figure}

\begin{figure}[!t]
\centering
\includegraphics[width=3.35in]{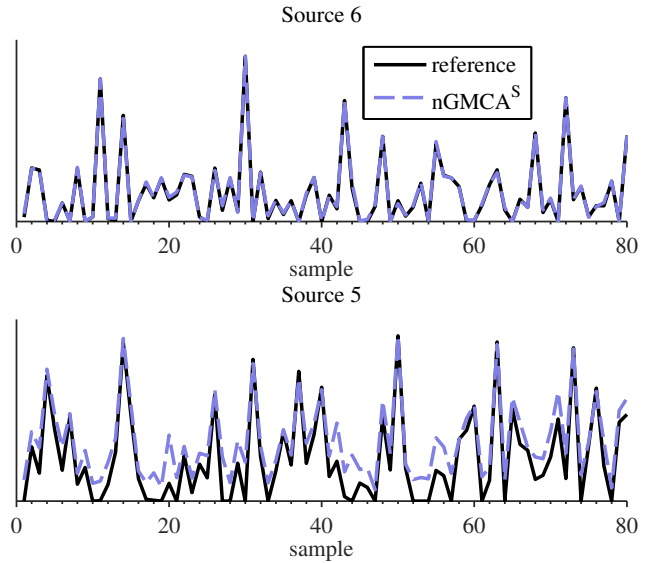}
\caption{Reference and estimated sources number 5 and 6\protect\\(nGMCA$^\text{S}$, $r=15$, $p_S=80\%$, noiseless, samples 1 to 80)}
\label{fig:biasedSources}
\end{figure}

With few sources, both nGMCA$^\text{S}$ and nGMCA$^\text{H}$ separate correctly as displays figure \ref{fig:r_a80n00}. The bias is therefore costly for nGMCA$^\text{S}$ since it leads to some compensation behaviors: with 80\% activation rate, nearly every coefficient suffers the bias and one source tends to compensate for all the others' with a positive offset. This can be seen on source 5 in figure \ref{fig:biasedSources}. The offset correlates with all the ground truth sources as can be observed from the correlation matrix in figure \ref{fig:correlationBias}: estimated source $5$ gathers all the thresholded coefficients of the other estimated sources and is therefore affected by the interferences.

On the other hand, the effect of soft-thresholding on the coefficients amplitude helps giving more weight to larger coefficients which is essential for the separation of sources from ill-conditioned mixtures as shown in figure \ref{fig:r_a80n00} with a large number of sources $r$; and in figure \ref{fig:Aalpha_r35a80} with a large $\alpha_A$ for instance.

\subsection{Noisy Data}
\label{sec:ExpNoisy}

In this section, noise is added to the data and the input of the algorithms is therefore $\bY=\bA^\text{ref}\bS^\text{ref}+\bs{Z}$, with $\bs{Z}$ a Gaussian matrix with independent and uniformly distributed coefficients. In the experiments, the amount of noise is given in term of SNR on the data $\bY$, SNR$_Y=10~\text{log}_{10}\left(\frac{\|\bA^\text{ref}\bS^\text{ref}\|_2^2}{\|\bY - \bA^\text{ref}\bS^\text{ref}\|_2^2}\right)$. nGMCA$^\text{naive}$, nGMCA$^\text{S}$ and nGMCA$^\text{H}$ are compared with Hoyer's \cite{Hoyer_04_Nonnegativematrix}, Kim \& Park's algorithms \cite{Kim_06_SparseNonnegativeMatrix} and sparse accelerated HALS \cite{Gillis_12_SparseandUnique}, which are competitive, publicly available algorithms and take sparsity into account in different ways (paragraph \ref{sec:sparseAlgorithms}). There is no straightforward way to set Kim \& Park's algorithm parameters and we therefore used the default parameters of the implementation. It is then left running until convergence. For Hoyer's algorithm, no prior is applied on $\bA$ and the sparsity ratio of $\bS$ is optimally tuned using the ground truth sources, since there is no automatic way to set the parameters. The sparsity level for the sparse accelerated HALS is also provided from the ground truth sources and both algorithms are left running for 5000 iterations in order to assure convergence. In all this section, $\tau_\sigma^\infty=1$ in the nGMCA algorithms, as an effective trade-off between noise removal and good separation of the sources.\medskip

The comparisons also include an oracle which solves the non-negatively constrained inversion problem in $\bS$  using the ground-truth sources $\bA^{\text{ref}}$:
\begin{equation}
\label{eq:oracle}
\underset{\bS\ge \b0}{\text{argmin}}~\frac{1}{2}\|\bY-\bA^{\text{ref}}\bS\|_2^2+\lambda \|\bS\|_1.
\end{equation}
The sparsity parameter is set to $\lambda = \tau_\sigma^\infty \sigma^\text{grad}$ such as in nGMCA$^\text{S}$. This oracle stands for the optimal $\bS$ which could be recovered by nGMCA$^\text{S}$ if the uncontaminated mixtures $A^{\text{ref}}$ were known. Of course, since the mixture are not known in practice in BSS, the oracle yields unachievable results, but it provides a reference line for the comparisons and a limit for the progression margin of the reconstructions.\medskip

\subsubsection{Summary of the experiments}

\begin{itemize}
\item Figures \ref{fig:noise_a10r15}, \ref{fig:noise_a10r15_oracle} and \ref{fig:noise_a30r15}: these benchmarks show the reconstruction results for 15 sources with activation rate of 10\% (figures \ref{fig:noise_a10r15} and \ref{fig:noise_a10r15_oracle}) and 30\% (figure \ref{fig:noise_a30r15}) ---the lower the activation rate, the better the sparse prior--- with a varying level of noise contamination in the data. Figure \ref{fig:noise_a10r15_oracle} and \ref{fig:noise_a30r15} display the loss in SDR compared to the oracle in order to facilitate the visualization.
In both cases, nGMCA$^\text{S}$ is less sensitive to noise and outperforms the other algorithms.

\begin{figure}[!t]
\centering
\includegraphics[width=3.35in]{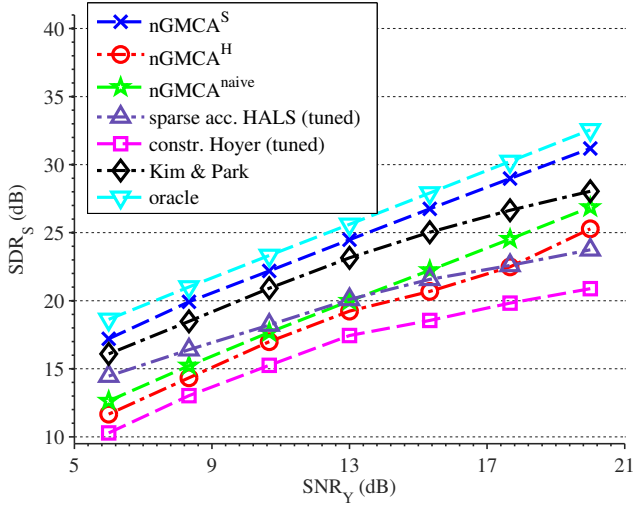}
\caption{Reconstruction SDR (SDR$_S$) with respect to the data SNR (SNR$_Y$) \protect\\($p_S=10\%$, $r=15$, average of 192 simulations)}
\label{fig:noise_a10r15}
\end{figure}

\begin{figure}[!t]
\centering
\includegraphics[width=3.35in]{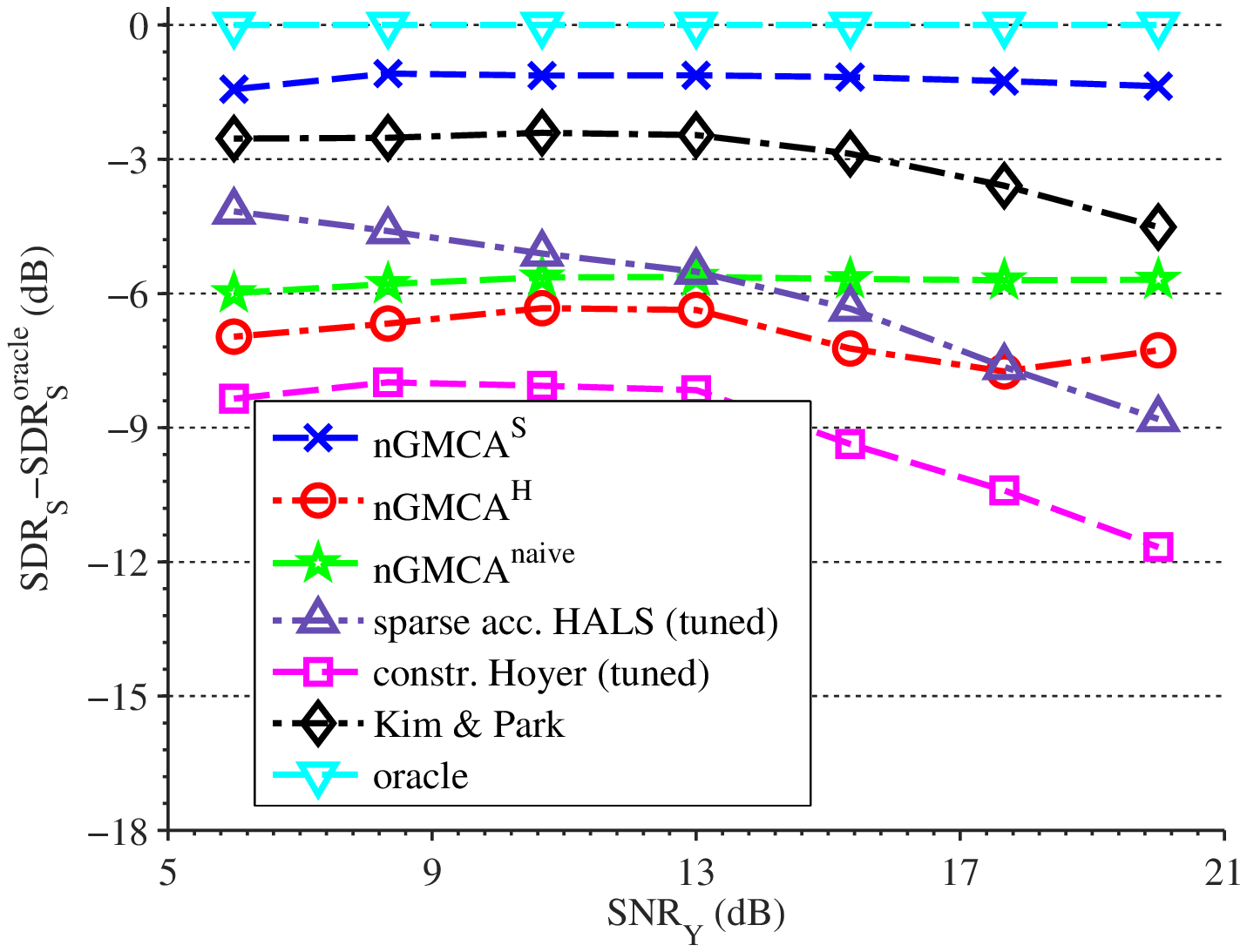}
\caption{Reconstruction SDR (SDR$_S$) minus the oracle SNR, with respect to the data SNR (SNR$_Y$) ($p_S=10\%$, $r=15$, average of 192 simulations)}
\label{fig:noise_a10r15_oracle}
\end{figure}

\begin{figure}[!t]
\centering
\includegraphics[width=3.35in]{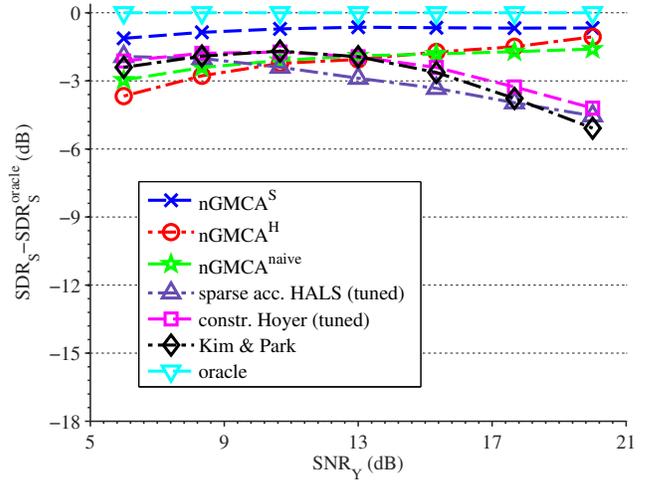}
\caption{Reconstruction SDR (SDR$_S$) minus the oracle SNR, with respect to the data SNR (SNR$_Y$) ($p_S=30\%$, $r=15$, average of 192 simulations)}
\label{fig:noise_a30r15}
\end{figure}

\smallskip
\item Figures \ref{fig:noise_a50r15}: this benchmark shows the same experience than the previous ones but with a larger activation rate (50\%) which is less favorable to the GMCA-based algorithms. nGMCA$^\text{S}$ remains better in most settings. Hoyer's algorithm performs similarly to nGMCA$^\text{S}$ for very noisy data but it is important to remember that in our experiment, Hoyer's algorithm and sparse accelerated HALS are provided with the ground truth sparsity ratios, which would not be available with such precision in practice. For cleaner data, nGMCA$^\text{H}$ and nGMCA$^\text{naive}$ begin to overtake nGMCA$^\text{S}$, which corroborates the results of the previous section for noiseless data with large activation rates and few sources (figure \ref{fig:r_a80n00}).\medskip

\begin{figure}[!t]
\centering
\includegraphics[width=3.35in]{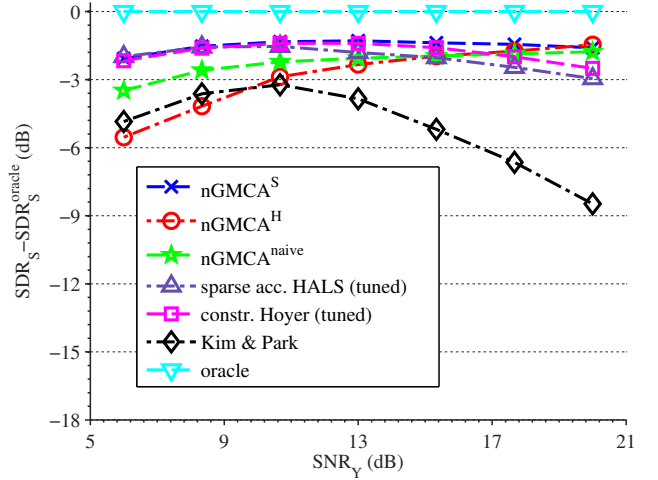}
\caption{Reconstruction SDR (SDR$_S$) minus the oracle SNR, with respect to the data SNR (SNR$_Y$) ($p_S=50\%$, $r=15$, average of 192 simulations)}
\label{fig:noise_a50r15}
\end{figure}

\smallskip
\item Figure \ref{fig:m_a30n03r15}: This benchmarks provides the reconstruction results for noisy data (15dB), 15 sources, a low activation rate (30\%) and a varying number of measurements $m$. The lower the number of measurements, the more difficult the reconstruction is, since the redundancy can help denoising and discriminating between the sources. While we have exhibited results for a large number of measurements so far, this shows that nGMCA$^\text{S}$ also compares favorably with other algorithms when the number of measurements is more restrained.\medskip

\begin{figure}[!t]
\centering
\includegraphics[width=3.35in]{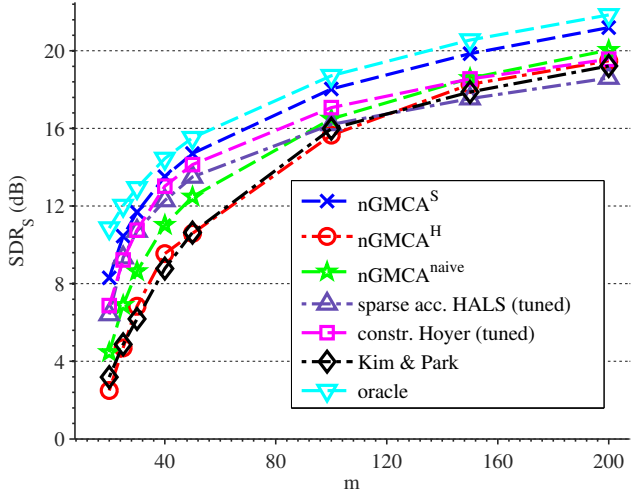}
\caption{Reconstruction SDR (SDR$_S$) with respect to the number of measurements $m$\protect\\(SNR$_Y$=15dB, $p_S=30\%$, $r=15$, average of 72 simulations)}
\label{fig:m_a30n03r15}
\end{figure}

\smallskip
\item Figure \ref{fig:r_15dBa30}: This benchmarks provides the reconstruction results for sparse (30\% activation rate) and noisy (15dB) data, and a varying number of sources $r$. The complexity of the separation rises with the number of sources hence the reconstruction results decrease with it for all algorithms, but in any case, nGMCA$^\text{S}$ performs best for all the values.\medskip

\begin{figure}[!t]
\centering
\includegraphics[width=3.35in]{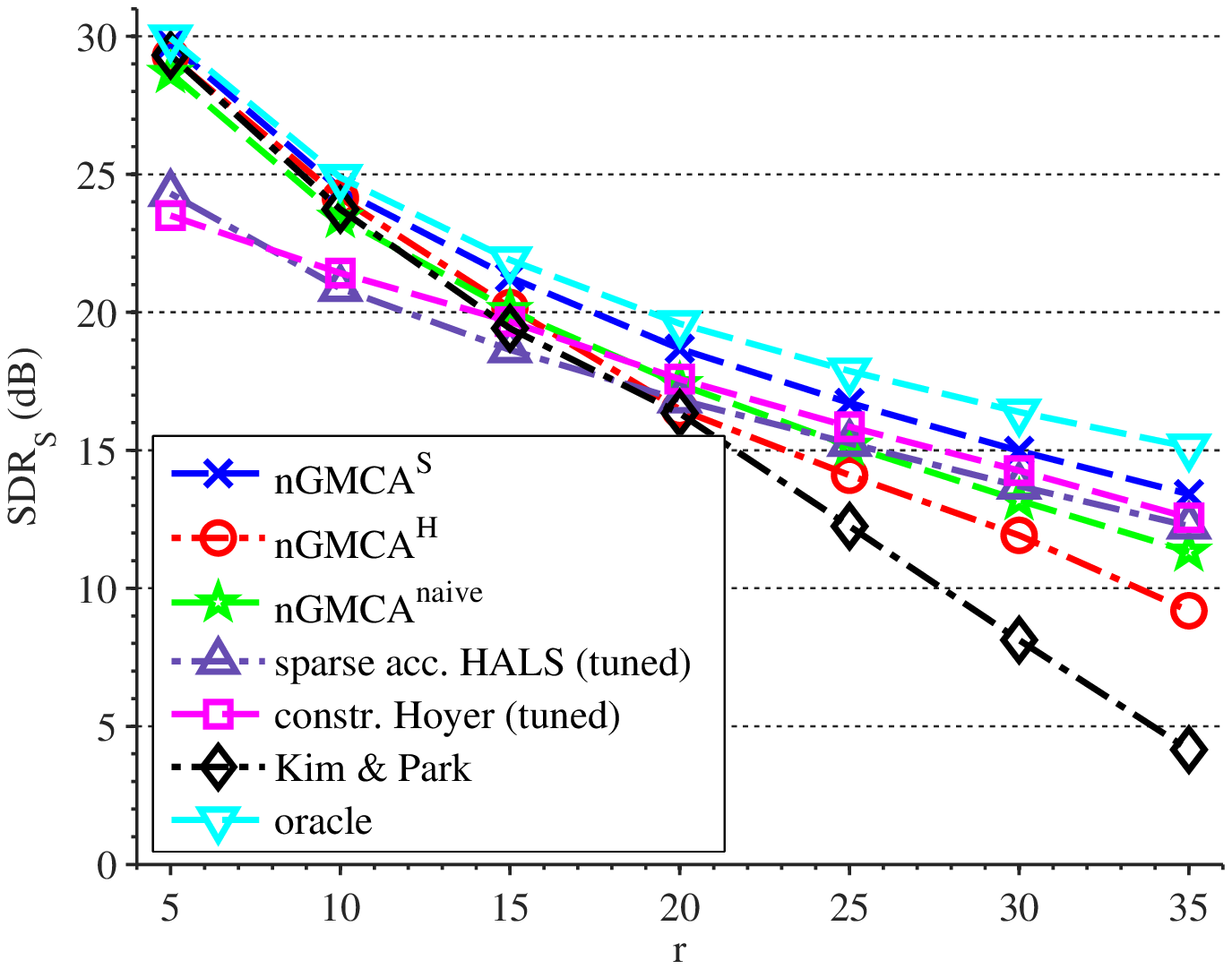}
\caption{Reconstruction SDR (SDR$_S$) with respect to the number of sources $r$ (SNR$_Y$=15dB, $p_S=30\%$, average of 48 simulations)}
\label{fig:r_15dBa30}
\end{figure}

\end{itemize}

\subsubsection{About the initialization and the separation}
\begin{figure}[!t]
\centering
\includegraphics[width=3.35in]{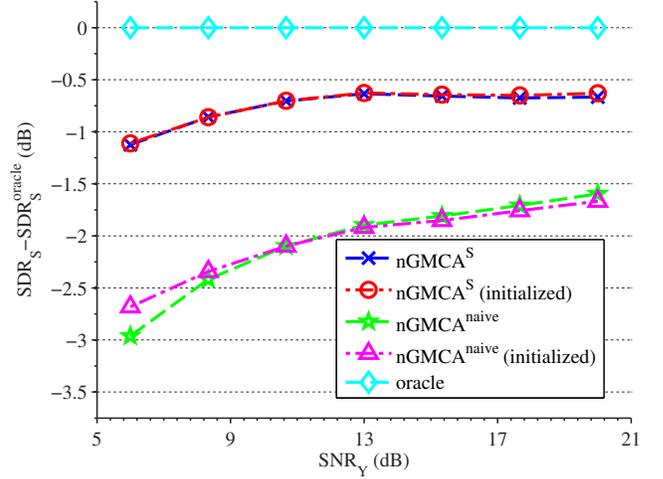}
\caption{Reconstruction SDR (SDR$_S$) with respect to the data SNR (SNR$_Y$)\protect\\($p_S=30\%$, $r=15$, average of 192 simulations)}
\label{fig:modelAndInit}
\end{figure}

~\\
Figure \ref{fig:modelAndInit} provides the same results as figure \ref{fig:noise_a30r15} but compares nGMCA$^\text{naive}$ and nGMCA$^\text{S}$ with version of them which are initialized with $\bA^\text{ref}$ and $\bS^\text{ref}$ and hence, with a perfect separation from the start. The initialized nGMCA$^\text{S}$ is also provided with the exact noise standard deviation. The difference in term of reconstruction quality between the regular algorithms and their optimally initialized version is extremely small. This shows that the automatic estimation of the noise level within nGMCA$^\text{S}$ is appropriate, and that the initialization of nGMCA$^\text{S}$ and nGMCA$^\text{naive}$ is robust. \medskip

\subsection{Conclusion of the Experiments - the Compromise}

Remember that the estimated sources (see \cite{Vincent_06_Performancemeasurementin}) can be decomposed as follows:
\begin{equation}
s^\text{est}=s_\text{target}+s_\text{interf}+s_\text{noise}+s_\text{artifacts},\label{eq:decomposition2}
\end{equation}
Any BSS algorithm must minimize at the same time interferences, noise and artifacts in order to achieve good performance. These three terms are strongly affected by the sparsity prior:
\begin{itemize}
\item \textit{Interferences}: They intervene when the sources are not correctly, or not completely, separated. The term $s_\text{interf}$ is computed as the projection on all the sources but the target. Sparsity, as a measure of diversity, can greatly help getting a correct separation of the sources, hence keeping this term relatively small. However, it can  still create interferences when the sparse model for the sources departs from their actual structure, such as in figure \ref{fig:biasedSources}. Interferences then originate from an imperfect source prior and/or badly separated sources.\\
\item \textit{Noise}: $s_\text{noise}$ is the part of the reconstruction that projects on the noise but not the sources. Since the Gaussian noise studied in this article spreads uniformly on all the coefficients, while sparse sources concentrate their energy on few coefficients, the thresholding effect implied by $\ell_0$ and $\ell_1$ regularizations significantly denoises the estimates. This reduces the importance of the noise term and therefore helps obtaining better reconstructions.\\
\item \textit{Artifacts}: for a given source, the artifacts $s_\text{artifacts}$ gathers the residues which are neither explained by the other sources nor the noise. We observed that the soft-thresholding operator introduces a bias which is the main contributor to the artifacts. Again, this term will increase when the sparsity level of the sources decreases: in such a case, the sparsity prior is not as well suited to constrain the morphology of the sources. It is also important to notice that even when the sources are very sparse, improperly constrained solutions are more akin to be contaminated with a higher level of artifacts.\\
\end{itemize}

All these  aspects interact with each other.  As shown in this section, nGMCA$^\text{S}$ provides an effective trade-off between noise, interferences and bias. Indeed, through the experiments, we show that nGMCA$^\text{S}$ outperforms other algorithms in most scenarios, according to the SDR criterion which takes into account these three origins of reconstruction deterioration.\\
For low activation rate (high sparsity), nGMCA$^\text{S}$ performs definitely better than the other algorithms for a large range of noise levels (figures \ref{fig:noise_a10r15} and \ref{fig:noise_a30r15}) while in the extreme noiseless case it performs quite reasonably. In this setting, the sparsity-enforcing $\ell_1$ prior plays its role at: i) getting a good separation process with respect to other priors (such as the $\ell_0$ pseudo-norm); this helps reducing the interferences, ii) correctly denoising the sources; this tends to lower the noise contribution and artifacts.

nGMCA$^\text{S}$ is noticeably quite robust to departures from the sparsity assumptions: it performs reasonably well with large activation rates (figures \ref{fig:r_a80n00} and \ref{fig:noise_a50r15}) but at the cost of a slight bias of the estimated sources (figures \ref{fig:correlationBias} and \ref{fig:biasedSources}) which tends to increase the contribution of the artifacts.

Additionally, the nGMCA$^\text{S}$ algorithm provides good separation performance for a large range of numbers of sources (figure \ref{fig:r_15dBa30}) as well as for ill-conditioned problems arising from a lack of observations in figure \ref{fig:m_a30n03r15}, or from correlated mixing directions in figure \ref{fig:Aalpha_r35a80}. These results can be explained by the good separation power of the $\ell_1$ regularizer with an appropriate tuning of the regularization parameter, in order to disentangle sparse sources, together with the appropriate implementation of the non-negativity constraints.

\section{Application}
\label{sec:Application}

\begin{figure}[!t]
\centering
\includegraphics[width=3.35in]{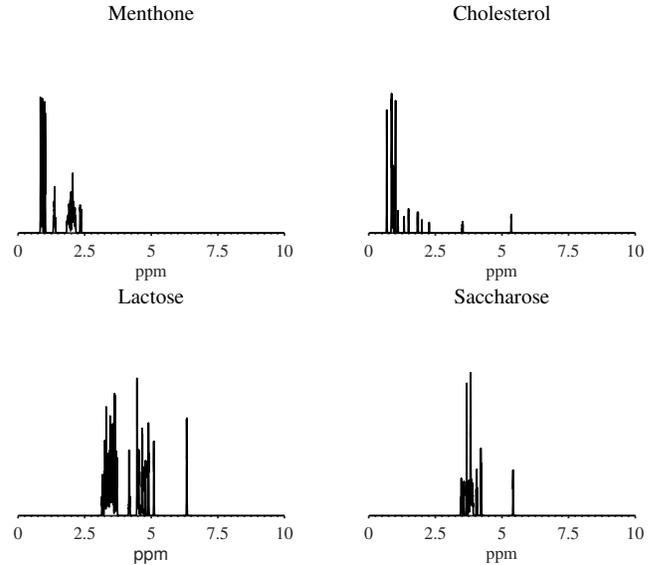}
\caption{NMR spectra of 4 chemical compounds.}
\label{fig:NMRspectra}
\end{figure}

\begin{figure}[!t]
\centering
\includegraphics[width=3.35in]{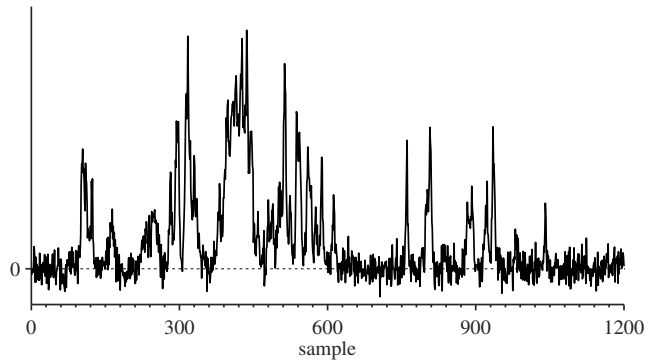}
\caption{Example of mixture (main component: lactose, SNR$_Y=15$dB)}
\label{fig:appli_mixture}
\end{figure}

\begin{figure}[!t]
\centering
\includegraphics[width=3.35in]{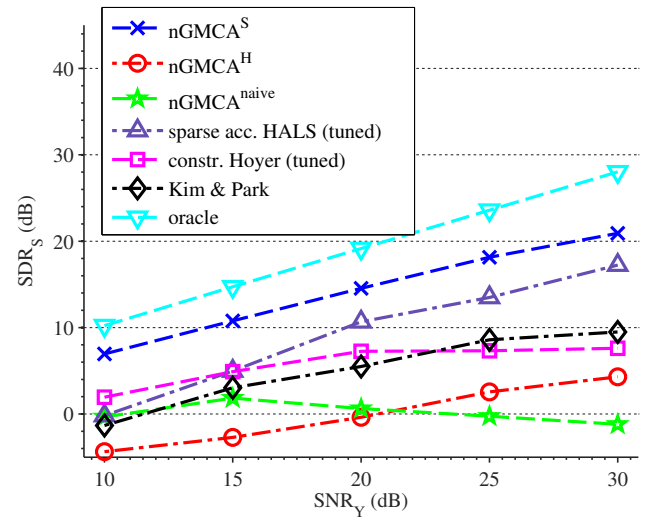}
\caption{Reconstruction SDR (SDR$_S$) with respect to the data SNR (SNR$_Y$) \protect\\($m=15$, $\tau_\sigma^\infty=2$, average of 96 synthetic NMR data simulations)}
\label{fig:noise_application}
\end{figure}

In physical applications, molecules can be identified by their specific Nuclear Magnetic Resonance (NMR) spectra. In this section, we simulate more realistic data, using NMR spectra of real molecules. These spectra are well adapted to the current settings since they are very sparse. The information about the peaks can be found in the Spectral Database for Organic Compounds, SDBS\footnote{\url{http://riodb01.ibase.aist.go.jp/sdbs/cgi-bin/cre_index.cgi}}. The spectra were convoluted with a Laplacian with width at half maximum of 3 samples, in order to account for the acquisition imperfections. The number of samples is set to $n=1200$. $\bS^\text{ref}$ is made of $r=15$ real spectra such as the ones displayed in figure \ref{fig:NMRspectra}. Some sources can exhibit strong normalized scalar product, such as cholesterol and menthone spectra for instance (0.67). The mixing coefficients of $\bA^\text{ref}$ are simulated in the same way as in the previous section ($p_A=1$, $\alpha_A=2$). The observed data is $\bY=\bA^\text{ref}\bS^\text{ref}+\bs{Z}$ where $\bs{Z}$ is an i.i.d. Gaussian noise matrix. An example of measurement where the lactose spectrum is particularly strong is provided in figure \ref{fig:appli_mixture}.\medskip

\begin{figure}[!t]
\centering
\includegraphics[width=3.35in]{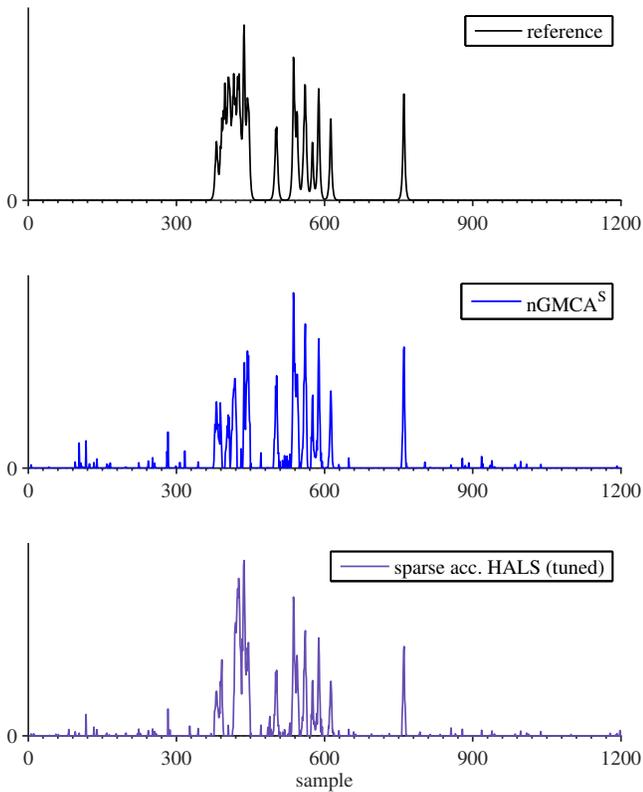}
\caption{Example of reconstruction (lactose, SNR$_Y=15$dB)}
\label{fig:appli_reconstruction_example}
\end{figure}

In figure \ref{fig:noise_application}, the number of measurements is limited to the number of sources, i.e. $m=15$, which occurs in some applications; and the curves show the influence of noise in the data. With so few measurements, denoising becomes more important, while at the same time the noise is underestimated by the algorithm since the problem is less constrained. To compensate this behavior, $\tau_\sigma^\infty$ is this time set to 2.

nGMCA$^\text{naive}$ fails to obtain suitable results. Indeed, in this setting the conditioning of the problem is extremely poor: $\text{cond}(\bA^T\bA)\approx 10^4$ and nGMCA$^\text{naive}$ is not able to converge. On the other hand, nGMCA$^\text{S}$ performs from 3 to 5dB better than all the other algorithms. This shows once again that nGMCA$^\text{S}$ is particularly robust for a large variety of settings. An example of reconstruction is given in figure \ref{fig:appli_reconstruction_example}, where nGMCA$^\text{S}$ is able to identify more peaks that sparse accelerated HALS. Its reconstruction is however not completely noiseless, since there is always a trade-off to find between denoising, separation and bias.\medskip

\begin{figure}[!t]
\centering
\includegraphics[width=3.35in]{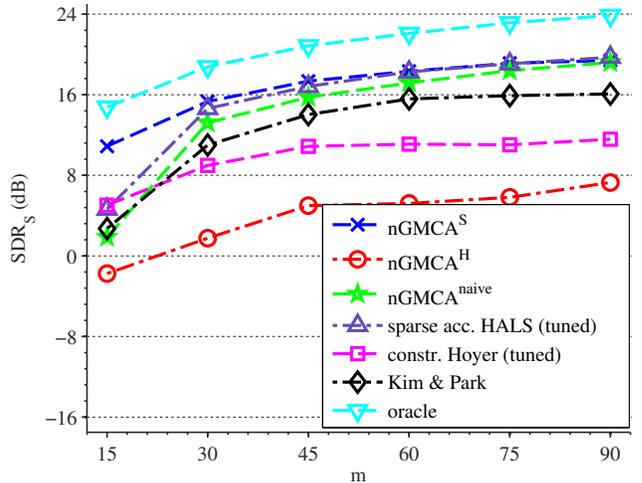}
\caption{Reconstruction SDR (SDR$_S$) with respect to the number of measurements $m$ (SNR$_Y=15$dB, $\tau_\sigma^\infty=2$, average of 36 synthetic NMR data simulations)}
\label{fig:m_application}
\end{figure}

In figure \ref{fig:m_application}, the number of measurements varies from 15 to 90. Since the conditioning greatly improves for larger numbers of measurements, nGMCA$^\text{naive}$ results increase very quickly. But in any case, although nGMCA$^\text{naive}$ and sparse accelerated HALS obtain similar results to nGMCA$^\text{S}$ when there are enough measurements, nGMCA$^\text{S}$ still performs better than all the other tested algorithms in most of the settings.

\section{Software}
\label{sec:software}
Following the philosophy of reproducible research \cite{Buckheit_95_WaveLabandReproducible}, the algorithms introduced in this article will be available at {\it http://www.cosmostat.org/GMCALab}.

\section{Conclusion}
In this paper we have introduced a new algorithm, nGMCA, to tackle the problem of sparse non-negative BSS from noisy mixtures. Inspired by a recent sparse BSS algorithm coined GMCA, several extensions have been explored which imply that a rigorous handling of both sparse and non-negative constraints are essential to avoid instabilities and sub-optimal solutions. In particular, one extension estimates both a mixing and a source matrix by exactly solving the non-negatively constrained and $\ell_1$ penalized sub-problems, using proximal techniques.
Extensive comparisons have been carried out with state-of-the-art algorithms on synthetic data; these experiments show that this nGMCA extension is robust to noise contamination thanks to a dedicated thresholding strategy, with negligible parameter tuning. The experiments also show that it performs well for a wide variety of settings, including problems with highly correlated mixture directions, few observations or a large number of sources. Finally, the nGMCA algorithm yields highly competitive results on synthetic mixtures of real NMR spectra.\medskip

In this article however, the sparsity of the sources only held in the direct or sample domain. Future work will focus on extending nGMCA to deal with the more general setting where the sources are still non-negative in the direct domain, but are sparse in a different signal representation.

\appendices

\section{Resolution of the sub-problems}
\label{app:FISTAsubproblem}
\textbf{Algorithm \ref{alg:FISTAsubS}} solves the sub-problem in $\bS$ \eqref{eq:subps} with $p=1$ using FISTA \cite{Beck_09_FastIterativeShrinkage}.

\begin{algorithm}
\caption{FISTA for the sub-problem in $\bS$}
\label{alg:FISTAsubS}
\begin{algorithmic}[1]
\Procedure{UpdateS}{$\bY,\bA,\lambda,\bS_0$}
\Require $\bY,\bA,\lambda$
\State \textbf{initialize} $\bs{R}_0=\bS_0$, $L=\|\bA^T\bA\|_{\text{s}}$, $t_1=1$, $k=1$
\While{not converged}

\State $\bS_k=\big[\text{Soft}_\frac{\lambda}{L}\big(\bs{R}_{k-1}-\dfrac{1}{L}\bA^T(\bA\bs{R}_{k-1}-\bY)\big)\big]_+$
\State $t_{k+1}=\dfrac{1+\sqrt{1+4t_k^2}}{2}$
\State $\bs{R}_k=\bS_k +\dfrac{t_k-1}{t_{k+1}} (\bS_k-\bS_{k-1})$
\State $k=k+1$
\EndWhile
\State \textbf{return} $\bS_k$
\EndProcedure
\end{algorithmic}
\end{algorithm}

\section*{Acknowledgment}
J.B. was supported by the French National Agency for Research (ANR) 11-ASTR-034-02-MultID. We thank the associate editor and the anonymous reviewers for their help in improving the clarity and quality of the paper. We also thank Daniel Machado for his help in writing the paper.

\ifCLASSOPTIONcaptionsoff
  \newpage
\fi

\end{document}